\documentclass[runningheads]{llncs}
\usepackage[T1]{fontenc}
\pdfoutput=1
\usepackage{graphicx}

\usepackage{hyperref}
\usepackage{color}

\usepackage{amsfonts}
\usepackage{amsmath}
\usepackage{bm}
\usepackage{bbm}
\usepackage{geometry}[margin=1in]
\usepackage{subcaption}
\usepackage{array}
\usepackage{multirow}

\usepackage[linesnumbered,ruled]{algorithm2e}

\usepackage{natbib}
\usepackage{algpseudocode}
\newcommand{\dfdx}{\frac{\partial f}{\partial x_s}}
\newcommand{\Rd}{\mathbb{R}^d}
\newcommand{\xb}{\mathbf{x}}
\newcommand{\xc}{\mathbf{x_c}}
\newcommand{\xcc}{\mathbf{x_{/s}}}

\newcommand{\Xcal}{\mathcal{X}}
\newcommand{\Ycal}{\mathcal{Y}}
\newcommand{\when}[1]{\mathbbm{1}_{#1}}

\begin{document}
\title{Regionally Additive Models: Explainable-by-design models minimizing feature interactions}

\titlerunning{Regionally Additive Models}

\author{
  Vasilis Gkolemis\inst{1, 2} % \orcidID{0000-0002-2636-0245}
  \and
  Anargiros Tzerefos\inst{1} % \orcidID{1111-2222-3333-4444}
  \and
  Theodore Dalamagas\inst{1} % \orcidID{2222--3333-4444-5555}
  \and
  Eirini Ntoutsi\inst{3} % \orcidID{2222--3333-4444-5555}
  \and
  Christos Diou\inst{2} % \orcidID{2222--3333-4444-5555}
}
\authorrunning{V. Gkolemis et al.}
% First names are abbreviated in the running head.
% If there are more than two authors, 'et al.' is used.
%
\institute{
   ATHENA RC \and
   Harokopio University of Athens \and
   Universitat der Bundeswehr Munchen
}
\maketitle              % typeset the header of the contribution
\begin{abstract}
Generalized Additive Models (GAMs) are widely used explainable-by-design models in various applications.
GAMs assume that the output can be represented as a sum of univariate functions, referred to as components.
However, this assumption fails in ML problems where the output depends on multiple features simultaneously.
In these cases, GAMs fail to capture the interaction terms of the underlying function, leading to subpar accuracy.
To (partially) address this issue, we propose Regionally Additive Models (RAMs), a novel class of explainable-by-design models.
RAMs identify subregions within the feature space where interactions are minimized.
Within these regions, it is more accurate to express the output as a sum of univariate functions (components).
Consequently, RAMs fit one component per subregion of each feature instead of one component per feature.
This approach yields a more expressive model compared to GAMs while retaining interpretability.
The RAM framework consists of three steps.
Firstly, we train a black-box model.
Secondly, using Regional Effect Plots, we identify subregions where the black-box model exhibits near-local additivity.
Lastly, we fit a GAM component for each identified subregion.
We validate the effectiveness of RAMs through experiments on both synthetic and real-world datasets.
The results confirm that RAMs offer improved expressiveness compared to GAMs while maintaining interpretability.
\end{abstract}
\section{Introduction}
\label{sec:intro}

% Paragraph for Motivating about Regionally Additive Models
Generalized Additive Models (GAMs)~\citep{hastie1987generalized} are a popular class of explainable by design
(x-by-design) models~\citep{rudin2019stop, ghassemi2021false}.
Their popularity stems from their inherent interpretability.
GAMs represent an aggregation of univariate functions, where the overall model can be expressed as
\(f(\xb) = c + \sum_{s=1}^D f_s(x_s)\).
Due to this structure, each individual univariate function (component) can be visualized and interpreted independently.
Consequently, understanding the behavior of the overall model simply requires visualizing all components, each with a one-dimensional plot.

However, GAMs have limitations especially in cases where the outcome depends on multiple features simultaneously, i.e., when the unknown predictive function includes terms that combine multiple features. Therefore, there are a lot of methods~\citep{enouen2022sparse, xu2023sparse, lou2013accurate, agarwal2021neural, chang2021node} that extend the traditional GAMs in multiple directions. The most famous direction involves selecting the most important higher-order interactions. GA$^2$Ms~\citep{lou2013accurate} first introduced this line of research extending the traditional GAMs by adding pairwise interactions in their
    formulation, i.e., \(f(\xb) = c + \sum_{s=1}^D f_s(x_s) + \sum_{s=1}^D \sum_{s_1 \neq s_2} f_{s_1 s_2}(x_{s_1}, x_{s_2})\).
GA$^2$Ms are also x-by-design, because the user can visualize both the first-order ($1D$ plots) and
second-order (($2D$ plots)) components.
As the number of features increases, the number of second-order interactions grows exponentially,
making it impractical for users to interpret a large number of two-dimensional plots.
Therefore, methods like GA$^2$Ms target on automatically selecting the most significant interaction terms.

Both GAMs and GA$^2$Ms have limitations in modeling interactions of more than two features,
and. The main reason behind this limitation is that it is difficult to visualize three or more features on a single plot. Therefore, an approach like that would violate the x-by-design principle.

To address this limitation, we propose a new class of x-by-design models called Regionally Additive Models (RAMs).
Since in the general case, it is infeasible to visualize terms with more than two variables, RAMs focus on learning terms with structure:
$f(x_{s_1} | \when{x_{c_1}}, \when{x_{c_2}}, \cdots)$ for first-degree interactions and
$f(x_{s_1}, x_{s_2} | \when{x_{c_1}}, \when{x_{c_2}}, \cdots)$ for second-degree interactions.
The symbol $\when{x_{c_1}}$ denotes the condition that the feature $x_{c_1}$ takes a specific value or belongs to a specific range.

To better grasp the idea, consider a prediction task where the outcome depends, among others,
on a combination $f(x_1, x_2, x_3)$ of three features:
$x_1 \in [20, 80]$ (age), $x_2 \in [0, 40]$ (years in work), and $x_3 \in \{True, False\}$ (married).
Both GAM and GA$^2$M would fail to accurately learn this term of the underlying predictive function.
However, the three-feature effect can be decomposed in two sets of second-degree conditional terms based
on the marital status: $f_{1}(x_1, x_2 | x_3 = True)$ and $f_{2}(x_1, x_2 | x_3 = False)$.
In this way, RAM can accurately represent $f$ through learning two second-degree conditional terms, one for each marital status.
Furthermore, the two sets of terms can be visualized and interpreted as using two-dimensional plots.
It is worth noting that the conditional terms can also include numerical features.
For example, it could be more accurate to learn instead a set of four first-degree terms, conditioned on the marital status and the years in work:
$f_{1}(x_1 | x_2 < 10, x_3 = True)$,
$f_{2}(x_1 | x_2 \geq 10, x_3 = True)$,
$f_{3}(x_1 | x_2 < 10, x_3 = False)$, and
$f_{4}(x_1 | x_2 \geq 10, x_3 = False)$, which can be visualized and interpreted as four one-dimensional plots.

To adhere to the x-by-design principle, RAMs should be able to automatically identify the
most significant conditional terms.
As the number of these terms increases, it becomes difficult for users to retain and interpret numerous plots
associated with each feature or pair of features.
Therefore, RAMs use Regional Effect Plots~\citep{herbinger2023decomposing} to identify a small set
of conditional terms that have the greatest impact in minimizing feature interactions.
The RAM framework consists of three key steps.
First, a black-box model is fitted to capture all high-order interactions.
Then, the subregions where the black-box model exhibits near-local additivity are identified using Regional Effect Plots.
Finally, a GAM component is fit to each identified subregion.

The main contributions of this paper are as follows:

\begin{itemize}
    \item We formulate a new class of x-by-design models called Regionally Additive Models (RAMs).
    \item We propose a generic framework for learning RAMs and we propose a novel method for identifying the most significant conditional terms.
    \item We demonstrate the effectiveness of RAMs in modeling high-order interactions on a synthetic toy example and two real-world datasets.
\end{itemize}

\section{Motivation}
\label{sec:motivation}

\begin{figure}[htbp]
    \centering
    \begin{subfigure}{0.32\textwidth}
        \centering
        \includegraphics[width=\textwidth]{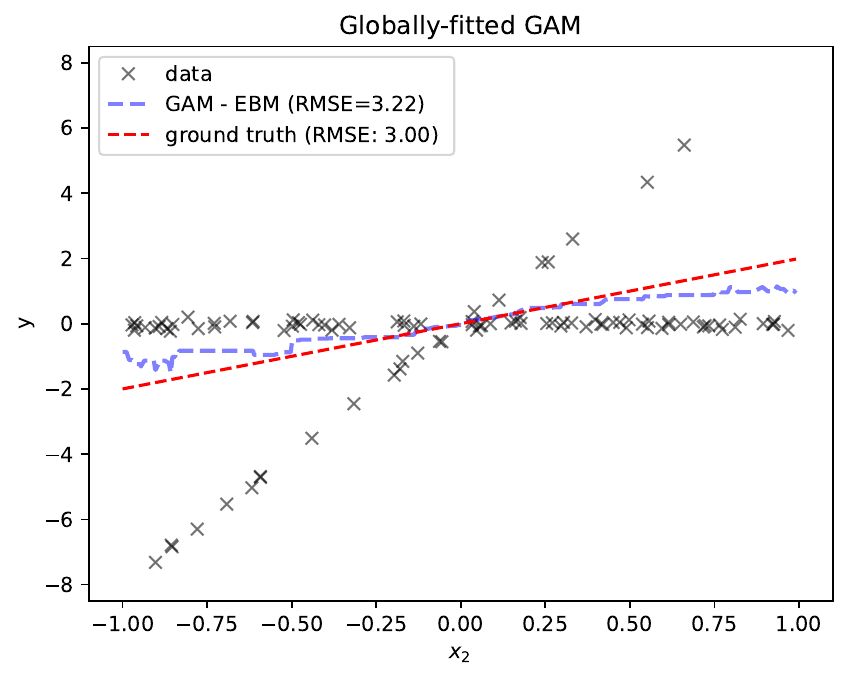}
        \caption{\(f_2(x_2)\)}
        \label{subfig:global_gam}
    \end{subfigure}
    \begin{subfigure}{0.32\textwidth}
        \centering
        \includegraphics[width=\textwidth]{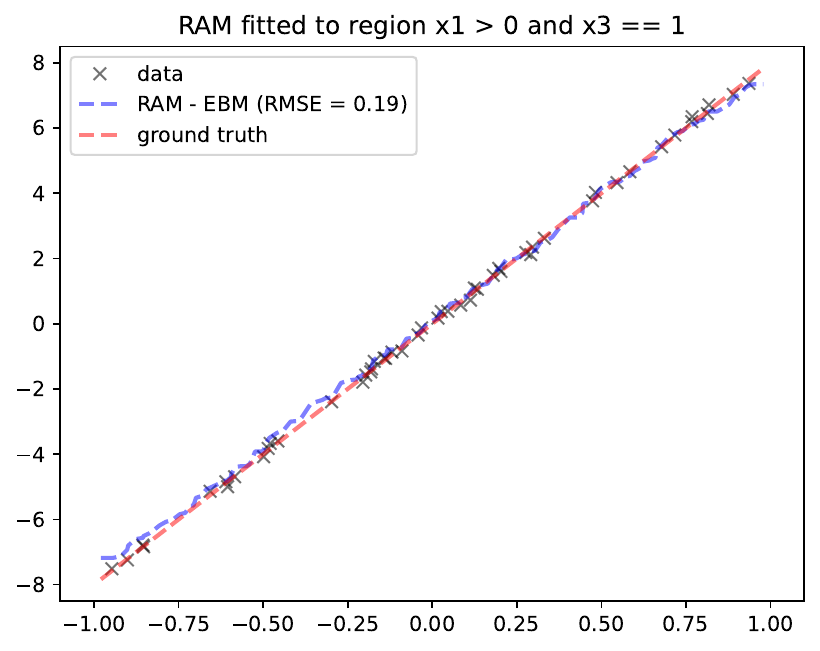}
        \caption{\(f_2(x_2) \when{x_1 > 0 \text{ and } x_3 = 1}\)}
        \label{subfig:regional_gam_1}
    \end{subfigure}
    \begin{subfigure}{0.32\textwidth}
        \centering
        \includegraphics[width=\textwidth]{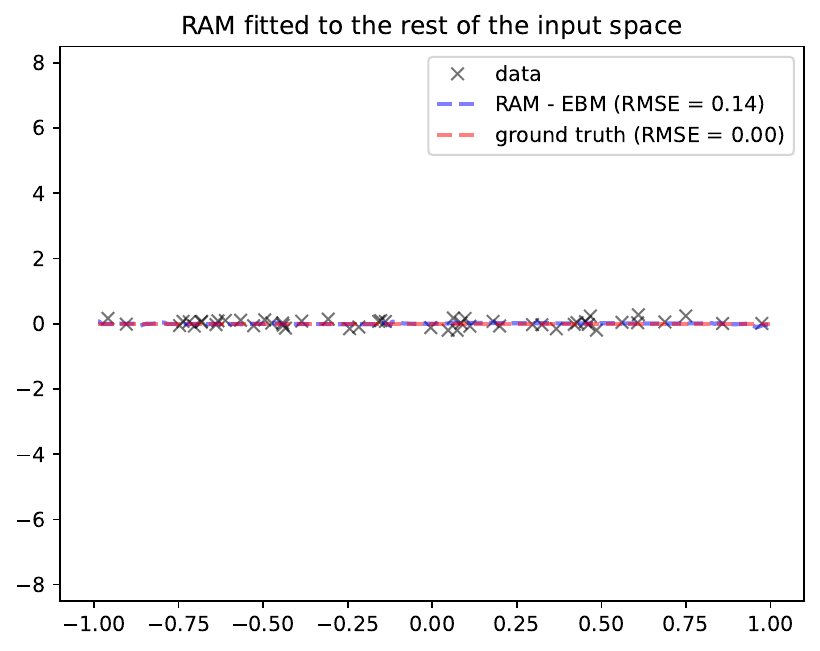}
        \caption{\(f_2(x_2) \when{x_1 \leq 0 \text{ or } x_3 \neq 1}\)}
        \label{subfig:regional_gam_2}
    \end{subfigure}
    \caption{The left image showcases the global GAM which erroneously learns an approximation of \(f(\xb) \approx 2x_2\).
    In contrast, the middle and right images demonstrate the RAM's ability to identify two distinct subregions where \(f\) exhibits regional additivity.
    By fitting a GAM to each subregion, the RAM accurately captures the true function $f$ while retaining interpretability.}
    \label{fig:ram_example}
\end{figure}

Consider the black-box function \(f(\xb) = 8x_2\when{x_1 > 0}\when{x_3=0}\)
with \(x_1, x_2 \sim \mathcal{U}(-1,1)\) and \(x_3 \sim Bernoulli(0,1)\).
Although very simple, GAM and GA$^2$M would fail to learn this mapping due to
the the three-features interaction term.
As we see in Figure~\ref{subfig:global_gam}, a GAM misleadingly learns that $\hat{f}(\xb) \approx 2x_2$,
because in $\frac{1}{4}$ of the cases ($x_1 > 0 \text{ and } x_3 = 0$) the impact of $x_2$ to the output is $8x_2$,
and in the rest $\frac{3}{4}$ of the cases the impact of $x_2$ to the output is $0$.
However, if splitting the input space in two subregions we observe that \(f\) is additive in each one (regionally additive):
\begin{equation}
    \label{eq:regionally_additive}
    f(\xb) = \begin{cases} 8x_2 & \text{if } x_1 > 0 \text{ and } x_3 = 1 \\ 0 & \text{otherwise} \end{cases}
\end{equation}
Therefore, if we knew the appropriate subregions,
namely, \(\mathcal{R}_{21} = \{x_1 > 0 \text{ and } x_3 = 0\}\)
and  \(\mathcal{R}_{22} = \{x_1 \leq 0 \text{ or } x_3 = 1\}\),
we could split the impact of $x_2$ appropriately and fit the following model to the data:

\begin{equation}
    \label{eq:regional_gam}
    f^{\mathtt{RAM}}(\xb) = f_1(x_1) + f_{21}(x_2) \when{(x_1, x_3) \in \mathcal{R}_{21}} + f_{22}(x_2) \when{(x_1, x_3) \in \mathcal{R}_{22}} + f_3(x_3)
\end{equation}
Equation~\eqref{eq:regional_gam} represents a Regionally Additive Model (RAM), which is simply a GAM fitted on each subregion of the feature space.
Importantly, RAM's enhanced expressiveness does not come at the expense of interpretability.
As we observe in Figures~\ref{subfig:regional_gam_1} and~\ref{subfig:regional_gam_2}, we can still visualize and comprehend each univariate function in isolation, exactly as we would do with a GAM,
with the only difference being that we have to consider the subregions where each univariate function is active,
The key challenge of RAMs is to appropriately identify the subregions where the black-box function is (close to) regionally additive.
For this purpose, as we will see in Section~\ref{subsec:regional_effect_methods}, we propose a novel algorithm that is based on the idea of
regional effect plots.

\section{RAM formulation}
\label{sec:ram_formulation}

\paragraph{Notation.}
Let \(\Xcal \in \Rd\) be the \(d\)-dimensional feature space, \(\Ycal\) the target space and \(f(\cdot) : \Xcal \rightarrow \Ycal\) the black-box function.
We use index \(s \in \{1, \ldots, d\}\) for the feature of interest and \(/s = \{1, \ldots, D\} - s\) for the rest.
For convenience, we use \((x_s, \xcc)\) to refer to \((x_1, \cdots , x_s, \cdots, x_D)\) and, equivalently, \((X_s, X_{/s})\) instead of \((X_1, \cdots , X_s, \cdots, X_D)\) when we refer to random variables.
The training set \(\mathcal{D} = \{(\xb^i, y^i)\}_{i=1}^N\) is sampled
i.i.d.\ from the distribution \(\mathbb{P}_{X,Y}\).
%Finally, \(f^{\mathtt{<method>}}(x_s)\) denotes how \(\mathtt{<method>}\)
%defines the feature effect and \(\hat{f}^{\mathtt{<method>}}(x_s)\)
%how it estimates it from the training set.

The RAM consists of a three-step pipeline; (a) fit a black-box model (Section~\ref{subsec:fit_black_box}),
(b) identify subregions with minimal interactions (Section~\ref{subsec:regional_effect_methods}) and
(c) fit a GAM component to each subregion (Section~\ref{subsec:fitting_gams}).

In step (b), we use regional effect methods~\citep{herbinger2023decomposing, herbinger2022repid}
to identify the regions where the black-box function is (close to) regionally additive.
% Describe the goal of the regional effect methods
Regional effect methods yield for each individual feature \(s\), a set of \(T_s\) non-overlapping regions,
denoted as \(\{\mathcal{R}_{st}\}_{t=1}^{T_s}\) where \(\mathcal{R}_{st} \subseteq \Xcal_{/s}\).
Note that, the number of non-overlapping regions can be different for each feature ($T_s$),
the regions \(\{\mathcal{R}_{st}\}_{t=1}^{T_s}\) are disjoint
and their union covers the entire feature space \(\Xcal_{/s}\).
The primary objective is to identify regions in which the impact of the \(s\)-th feature on the output is
\textit{relatively independent} of the values of the other features \(\xcc\).
To better grasp this objective, if we decompose the impact of the \(s\)-th feature on the output $y$ into two terms:
\(f_s(x_s, \xcc) = f_{s,ind}(x_s) + f_{s, int}(x_s, \xcc)\),
where \(f_{s,ind}(\cdot)\) represents the independent effect
and \(f_{s, int}(\cdot)\) represents the interaction effect,
the objective is to identify regions \(\{\mathcal{R}_{st}\}_{t=1}^{T_s}\) such that the interaction effect is minimized.
Regionally Additive Models (RAM) formulate the mapping \(\mathcal{X} \rightarrow \mathcal{Y}\) as:

\begin{equation}
\label{eq:ram_formulation}
f^{\mathtt{RAM}}(\xb) = c + \sum_{s=1}^D \sum_{t=1}^{T_s} f_{st}(x_s) \when{\xcc \in \mathcal{R}_{st}}, \quad \xb \in \Xcal
\end{equation}
In the above formulation, \(f_{st}(\cdot)\) is the component of the \(s\)-th feature which is active on the \(t\)-th region.
RAM can be viewed as a GAM with \(T_s\) components per feature where each component is applied to a specific region \(\mathcal{R}_{st}\).
To facilitate this interpretation, we can define an enhanced feature space \(\Xcal^\mathtt{RAM}\) defined as:

\begin{equation}
\label{eq:ram_feature_space}
\begin{aligned}
\Xcal^{\mathtt{RAM}} &= \{x_{st} | s \in \{1, \ldots, D\}, t \in \{1, \ldots, T_s\}\} \\
x_{sk} &= \begin{cases}
x_s, & \text{if } \xcc \in \mathcal{R}_{sk} \\
0, & \text{otherwise}
\end{cases}
\end{aligned}
\end{equation}
and then define RAM as a typical GAM on the extended feature space \(\Xcal^{\mathtt{RAM}}\):

\begin{equation}
\label{eq:ram_formulation2}
f^{\mathtt{RAM}}(\xb) = c + \sum_{s,t} f_{st}(x_{st}) \quad \xb \in \Xcal^{\mathtt{RAM}}
\end{equation}
Equations~\ref{eq:ram_formulation} and~\ref{eq:ram_formulation2} are equivalent.
To better understand of the formulations, consider the toy example described in Section~\ref{sec:motivation}.
To minimize the impact of feature interactions, we need to divide feature \(x_2\) into two subregions,
\(\mathcal{R}_{21} = \{x_1 > 0 \text{ and } x_3 = 1\}\) and \(\mathcal{R}_{22} = \{x_1 \leq 0 \text{ or } x_3 = 0\}\).
Using Eq.~\ref{eq:ram_formulation}, RAM formulation is:
\(f^{\mathtt{RAM}}(\xb) = f_1(x_1) + f_{21}(x_2) \when{x_1 > 0 \text{ and } x_3 = 1} + f_{22}(x_2) \when{x_1 \leq 0 \text{ or } x_3 = 0} + f_3(x_3)\).
Using Eq.~\ref{eq:ram_feature_space}, we should first define the augmented feature space
\(\Xcal^{\mathtt{RAM}} = (x_1, x_{21}, x_{22}, x_3)\),
where \(x_{21} = x_2 \when{x_1 > 0 \text{ and } x_3 = 1}\) and \(x_{22} = x_2 \when{x_1 \leq 0 \text{ or } x_3 = 0}\)
and then RAM formulation is: \(f^{\mathtt{RAM}}(\xb) = f_1(x_1) + f_{21}(x_{21}) + f_{22}(x_{22}) + f_3(x_3)\).

\section{RAM framework}
\label{sec:ram_framework}

\subsection{First step: Fit a black-box function}
\label{subsec:fit_black_box}

In the initial step of the pipeline, we fit a black-box function \(f(\cdot)\) to the training set
\(\mathcal{D} = \{(\xb^i, y^i)\}_{i=1}^N\) to accurately learn the underlying mapping \(f(\cdot) : \Xcal \rightarrow \Ycal\).
While any black-box function can theoretically be employed in this stage,
for utilizing the DALE approximation, as we will show in the next step,
it is necessary to select a differentiable function.
Recent advancements have demonstrated that differentiable Deep Learning models,
specifically designed for tabular data~\citep{arik2021tabnet}, are capable of achieving state-of-the-art performance,
making them a suitable choice for this step.
% Driver, teacher, guide

\subsection{Second step: Find subregions}
\label{subsec:regional_effect_methods}

To identify the regions of the input space where the impact of feature interactions is reduced,
we have developed a regional effect method influenced by the research conducted by
\citet{herbinger2023decomposing} and \citet{gkolemis2023dale}.
\citet{herbinger2023decomposing} introduced a versatile framework for detecting such regions,
where one of the proposed methods is the Accumulated Local Effects~\citep{apley2020visualizing}.
We have adopted their approach with two notable modifications.
First, instead of using the ALE plot, we employ the Differential ALE (DALE) method introduced by \citet{gkolemis2023dale},
which provides considerable computational advantages when the underlying black-box function is differentiable.
Second, we utilize variable-size bins, instead of the fixed-size ones in DALE, because the result in a more accurate approximation, as show by \citet{gkolemis2023rhale}.

\paragraph{DALE}

DALE gets as input the black-box function \(f(\cdot)\)
and the dataset \(\mathcal{D} = \{(\xb^i, y^i)\}_{i=1}^N\),
and returns the effect (impact) of the $s$-th feature $s$ on the output $y$:
\begin{equation}  \label{eq:DALE_accumulated_mean_est}
  \hat{f}^{\mathtt{DALE}}(x_s) = \Delta x \sum_{k=1}^{k_x} \underbrace{\frac{1}{|\mathcal{S}_k|} \sum_{i:\mathbf{x}^{(i)} \in
    \mathcal{S}_k} \dfdx(\mathbf{x}^i)}_{\hat{\mu}(z_{k-1}, z_k)})
\end{equation}
For more details on the DALE method, please refer to the original paper~\citep{gkolemis2023dale}.
In the above equation, \(k_x\) is the index of the bin such that
\(z_{k_x-1} \leq x_s < z_{k_x} \) and \(\mathcal{S}_k\)
is the set of the instances of the \(k\)-th bin, i.e.
\( \mathcal{S}_k = \{ \xb^i : z_{k-1} \leq x^{(i)}_s < z_{k} \} \).
In short, DALE computes the average effect (impact) of the feature \(x_s\) on the output,
by, first, dividing the feature space into $K$ equally-sized bins, i.e., \(z_0, \ldots, z_K\)
second, computing the average effect in each bin \(\hat{\mu}(z_{k-1}, z_k)\) (bin-effect) as the average of the instance-level effects inside the bin,
and, finally, aggregating the bin-level effects.

\paragraph{DALE for feature interactions}

In cases where there are strong interactions between the features,
the instance-level effects inside each bin deviate from the average bin-effect (bin-deviation).
We can measure such deviation using the standard deviation of the instance-level effects inside each bin:

\begin{equation}
  \label{eq:var_bin_approx}
  \hat{\sigma}^2(z_{k-1}, z_k) = \frac{1}{|\mathcal{S}_k| - 1}
\sum_{i:\mathbf{x}^i \in \mathcal{S}_k} \left ( \dfdx(\mathbf{x}^i) -
  \hat{\mu}(z_{k-1}, z_k) \right )^2
\end{equation}
The bin-deviation is a measure of the interaction between the feature \(x_s\) and the rest of the features inside the \(k\)-th bin.
Therefore, we can measure the global interaction between the feature \(x_s\) and the rest of the features along the whole \(s\)-th dimension
with the aggregated bin-deviation:
\begin{equation}
  \label{eq:DALE_interaction}
  \mathcal{H}_s = \sqrt{ \sum_{k=1}^{k_x} (z_k - z_{k-1})^2 \hat{\sigma}^2(z_{k-1}, z_k) }
\end{equation}
Eq.~\eqref{eq:DALE_interaction} outputs values in the range \([0, \infty)\) with zero indicating that \(x_s\) does not interact with any other feature,
i.e., the underlying black box function can be written as $f(\xb) = f_s(x_s) + f_{/s}(x_{/s})$.
In all other cases, $\mathcal{H}_s$ is greater than zero and the higher the value, the stronger the interaction.

A final detail, is that in order to have a more robust estimation of the bin-effect and the bin-deviation,
we use variable-size bins instead of the fixed-size ones in DALE.
In particular, we start with a dense fixed-size grid of bins and we iteratively merge the neighboring bins with similar
bin-effect and bin-deviation until all bins have at least a minimum number of instances.
In this way, we can have a more accurate approximation of the bin-effect and the bin-deviation.

\paragraph{Subregions as an optimization problem}

In the same way that we can estimate the feature effect (Eq.~\eqref{eq:DALE_accumulated_mean_est}) and
the feature interactions (Eq.~\eqref{eq:DALE_interaction}) for the $s$-th feature in the whole input space,
we can also estimate the effect and the interactions in a subregion of the input space \(\mathcal{R}_{st} \subset \mathcal{X}\).
We denote the equivalent regional qunatities as $\hat{f}^{\mathtt{DALE}}_{st}(x_s)$ and $\mathcal{H}_{st}$.
$f^{\mathtt{DALE}}_{\mathcal{R}_{st}}(x_s)$ and $\mathcal{H}_{\mathcal{R}_{st}}$ are defined exactly as in
Eq.~\eqref{eq:DALE_accumulated_mean_est} and Eq.~\eqref{eq:DALE_interaction} respectively,
with the only difference that instead of using the whole dataset \(\mathcal{D}\), to compute the regional bin-effect $\hat{\mu}_{st}(z_{k-1}, z_k)$
and the regional bin-deviation $\hat{\sigma}_{st}^2(z_{k-1}, z_k)$,
we use $\mathcal{D}_{st}$ which includes only the instances that belong to the subregion \(\mathcal{R}_{st}\),
i.e., $\mathcal{D}_{st} = \{\xb^i: x_s^i \in \mathcal{S}_k \land \xc^i \in \mathcal{R}_{st}\}$.
Therefore, in order to minimize the interactions of a particular feature $s$ we search for a set of regions \(\{\mathcal{R}_{st}\}_{t=1}^{T_s}\),
that minimizes the following objective:

\begin{equation}
  \label{eq:optimal_subregions}
  \begin{aligned}
    & \underset{\{\mathcal{R}_{st}\}_{t=1}^{T_s}}{\text{minimize}}
    & & \mathcal{L}_s = \sum_{t=1}^{T_s} \frac{| \mathcal{D}_{st} |}{|\mathcal{D}|} \mathcal{H}_{st} \\
    & \text{subject to}
    & & \bigcup_{t=1}^{T} \mathcal{R}_{st} = \mathcal{X} \\
    & & & \mathcal{R}_{st} \cap \mathcal{R}_{s\tau} = \emptyset, \quad \forall t \neq \tau
  \end{aligned}
\end{equation}
In Eq.~\eqref{eq:optimal_subregions}, the objective function is the weighted sum of the regional interactions $\mathcal{H}_{st}$,
where the weights are the number of instances in each subregion.
In this way, we give more importance to the subregions that contain more instances.
The first constraint ensures that the subregions cover the whole input space and the second constraint ensures that the subregions are disjoint.

\paragraph{Proposed solution}

The core of the method is outlined in Algorithm~\ref{alg:ram}.
First, we fit a differentiable black box model to the data (Step 1) and we copute the Jacobian matrix w.r.t. the input features (Step 2).
Then we search for a set of subregions by minimizing the objective of Eq.~\eqref{eq:optimal_subregions}
for each feature $s$ independently (Steps 3-4-5).
Based on the optimal subregions, we define the extended feature space (Step 6) and we fit a GAM in the extended feature space (Step 7).

For solving Eq.~\eqref{eq:optimal_subregions}, we have developed a tree-based algorithm
based on the approach proposed by~\citep{herbinger2023decomposing}, which we describe in detail in Algorithm~\ref{alg:subregion_detection}.
To describe the algorithm, we define some additional notation: $\mathcal{R}_s^l$ is the set of optimal subregion of the $s$-th feature at level $l$ of the tree.
Since at each level of the tree we divide the input space into two subregions, at level $l$ we have $2^l$ subregions,
i.e., $\mathcal{R}_s^l = \{\mathcal{R}_{st}\}_{t=1}^{2^l}$.
Equivalently, $\mathcal{L}_s^l$ is the optimal objective value of Eq.~\eqref{eq:optimal_subregions} at level $l$ of the tree.
Although the algorithm can search for an abritrary number of subregions per feature,
in order to preserve the smooth interpretation of the method,
we limit the maximum depth of the tree to $L=3$ levels,
which stands for a maximum of $T = 2^L = 8$ subregions per feature.
In general, the user can control the trade-off between the interpretability and the accuracy of the method by changing the maximum depth of the tree.
Note that with three splits, we already have an interaction term of
four ($f(x_s | \when{x_{c_1}}, \when{x_{c_2}}, \when{x_{c_3}}$) or
five ($f(x_{s_1}, x_{s_{2}} | \when{x_{c_1}}, \when{x_{c_2}}, \when{x_{c_3}}$) features.

To describe how the algorithm finds the optimal splits at each level $l$, let's consider the illustrative example of Section~\ref{sec:motivation}.
For feature $s=2$, the algorithm starts with $/s = \{1, 3\}$ as candidate split-features for the first level of the tree.
For each candidate split-feature, the algorithm determines the candidate split positions.
Since $x_1$ is a continuous feature, the candidate splits postions are a linearly spaced grid of $P$ points within the range of the feature, i.e. $[-1, 1]$,
where $P$ is a hyperparameter of the algorithm, set to $10$ in the experiments.
Therefore, the candidate positions are $p \in \{-1, -0.8, -0.6, \ldots, 0.8, 1\}$ each on defining two subregions,
$\mathcal{R}_{21} = \{ (x_1, x_3) : x_1 \leq p \}$ and $\mathcal{R}_{22} = \{ (x_1, x_3) : x_1 > p \}$.
As for $x_3$, being a categorical feature, the candidate split points are its unique values, i.e., $\{0, 1\}$,
and the corresponding subregions are $\mathcal{R}_{21} = \{ (x_1, x_3) : x_3 = 0 \}$ and $\mathcal{R}_{22} = \{ (x_1, x_3) : x_3 \neq 0 \}$.
Each candidate position, creates a corresponding dataset $[\mathcal{D}_{21}, \mathcal{D}_{22}]$,
and the algorithm computes the weighted level of interactions $\mathcal{H}_{21}$ and $\mathcal{H}_{22}$ for each dataset.
After iterating over all features and all candidate positions for each feature,
it selects the split point that minimizes the weighted level of interactions.
In the illustrative example, the optimal first-level split is based on $x_3$ and the optimal split point is $p = 0$.
The algorithm next proceeds to the second level, where the only candidate feature is $x_3$.
In this step, the first split is considered fixed so the optimal second split is applied to the subregions $\mathcal{R}_{21}$ and $\mathcal{R}_{22}$,
creating four subregions in total.
The algorithm continues in a similar manner, until it reaches the maximum depth $T$ or the drop in the weighted level of interactions is below a threshold $\epsilon$ (set to $20\%$ drop in the experiments).

\begin{algorithm}
\caption{Regionally Additive Model (RAM) training}
\label{alg:ram}
\SetAlgoLined
\SetKwInOut{Input}{Input}
\SetKwInOut{Output}{Output}
\BlankLine
\Input{A dataset $(X, y)$ and a maximum level $T$}
\Output{A trained RAM model $f^{\mathtt{RAM}}$}
  \BlankLine
  Train a differentiable black box model $f$ using $(X, y)$\;
  Compute the Jacobian w.r.t. features $\xb$,  $J = \nabla_{\xb}f(\xb)$ \;
  \For{$s \in \{1, \ldots, D\}$}{
    $\left\{\mathcal{R}_{st}\right\}_{t=1}^{T_s}$ = DetectSubregions($X$, $J$, $T$, $s$)\;
  }
  Create the extended feature space $\mathcal{X}^{\mathtt{RAM}}$ using all
  $\mathcal{R}_{st}$, as in Eq. \eqref{eq:ram_feature_space} \;
  Fit a GAM in $\mathcal{X}^{\mathtt{RAM}}$ \tcp*{i.e., train each $f_{st}$ using
    only data in $\mathcal{R}_{st}$}
  \Return{$f^{\mathtt{RAM}}(\xb) = c + \sum_{s,t} f_{st}(x_{st}), \quad \xb \in \Xcal^{\mathtt{RAM}}$}
\end{algorithm}

\begin{algorithm}
\caption{DetectSubregions}
\label{alg:subregion_detection}
\SetAlgoLined
\SetKwInOut{Input}{Input}
\SetKwInOut{Output}{Output}
\BlankLine
\Input{Dataset $X$, Gradients $J$, Maximum depth $T$, Feature $s$}
\BlankLine
\Output{Subregions $\{\mathcal{R}_{st}\}_{t=1}^{T_s}$, where $T_s \leq 2^T$}
\BlankLine
$\mathcal{H}_s^0$ \tcp*{Compute the level of interactions before any split}
$T_s = 0$ \tcp*{Initialize the number of splits for feature $s$}
\For{$l = 1$ to $L$}{
    \If{$H_{s}^{l-1} = 0$}{
        break\;
    }
    \tcc{Find best split feature $c_s^l$ at point $p_s^l$, leading to loss
        $\mathcal{H}_s^l$ using regions of previous level}
    Find $\mathcal{H}_{st}, c, p$ of the optimal split based on $\mathcal{R}_s^l$ \;
    \If{$1 - \frac{\mathcal{H}_s^l}{\mathcal{H}_s^{l-1}} > \epsilon$}{
        break\;
    }
    $T_s = 2^l$ \tcp*{Update the number of splits for feature $s$}
}
\Return{$\left\{\mathcal{R}_{st}|s\in\left\{1,\ldots,D\right\},
  t\in\left\{1,\ldots,T_s\right\}\right\}$}
\end{algorithm}

\paragraph{Computational Complexity}

Algorithm~\ref{alg:subregion_detection} has a computational complexity of $\mathcal{O}(D-1 \cdot L \cdot N)$
as it iterates over all features, query positions, and performs indexing operations on the data (splitting the dataset and computing the level of interactions).
Algorithm~\ref{alg:subregion_detection} is applied to each feature $s$ independently, and so computational complexity of the entire algorithm is $\mathcal{O}(D \cdot (D-1) \cdot L \cdot N)$.
However, in practice, $P$ and $T$ are small numbers.
Therefore, the computational complexity of the proposed method simplifies to $\mathcal{O}(D^2 \cdot N)$,
making it suitable for large datasets, heavy models, and reasonably high-dimensional data.
The key point is that the use of DALE eliminates the need to compute the Jacobian matrix for each split,
which is the most computationally expensive step.
This is because the Jacobian matrix is computed only once for the entire dataset, and then it is used as a lookup table for computing the level of interactions for each split.
This makes the proposed method applicable to heavy models.

\subsection{Third step: Fit a GAM in each subregion}
\label{subsec:fitting_gams}

Once the subregions are detected, any Generalized Additive Model (GAM) family can be fitted to the augmented input space $\mathcal{X}^{\mathtt{RAM}}$.
Recently, several methods have been proposed to extend GAMs and enhance their expressiveness.
These methods can be categorized into two main research directions.
The first direction targets on representing the main components of a GAM $\{ f_i(x_i) \}$ using novel models.
For example,~\citep{agarwal2021neural} introduced an approach that employs an end-to-end neural network to learn the main components.
The second direction aims to extend GAMs to model feature interactions.
Examples of such extensions include Explainable Boosting Machines (EBMs)~\citep{lou2013accurate} or
Node-GAMs~\citep{chang2021node}.
These models are generalized additive models that incorporate pairwise interaction terms.
It is worth noticing, that the RAM framework and can be used on top of both these research directions
to further enhance the expressiveness of the models while maintaining their interpretability.
In our experiments, we use the Explainable Boosting Machines (EBMs).

\section{Experiments}

We evaluate the proposed approach on two typical tabular datasets:
the Bike-Sharing Dataset~\citep{misc_bike_sharing_dataset_275} and the California Housing Dataset~\citep{pace1997sparse}.

\begin{table}[htbp]
  \centering
  \caption{The table compares the Mean Absolute Error (MAE) and the Root Mean Square Error (RMSE) of
  DNN, GAM, RAM, GA$^2$M, and RA$^2$M (representing 2nd order interactions), on two datasets: Bike-Sharing and California Housing.
  Lower values indicate better performance.
  RAM consistently outperforms GAM and approaches DNN performance.}
  \label{tab:sample}
  \begin{tabular}{l|c|cccc}
      & \textbf{Black-box} & \multicolumn{4}{c}{\textbf{x-by-design}} \\
      \hline
      \hline
      & all orders & \multicolumn{2}{c}{1\textsuperscript{st} order} & \multicolumn{2}{c}{2\textsuperscript{nd} order} \\
      \hline
      \hline
      & \textbf{DNN} & \textbf{GAM} & \textbf{RAM} & \textbf{GA}$^2$\textbf{M} & \textbf{RA}$^2$\textbf{M} \\
      \hline
      Bike Sharing (MAE)  & 0.254 & 0.549 & 0.430 & 0.298 & 0.278 \\
      Bike Sharing (RMSE) & 0.389 & 0.734 & 0.563 & 0.438 & 0.412 \\
      \hline
      California Housing (MAE)  & 0.373 & 0.600 & 0.553 & 0.554 & 0.533 \\
      California Housing (RMSE) & 0.533 & 0.819 & 0.754 & 0.774 & 0.739 \\
  \end{tabular}
\end{table}

\paragraph{Bike-Sharing Dataset}

The Bike-Sharing dataset contains the hourly bike rentals in the state of Washington DC over the period 2011 and 2012.
The dataset contains a total of 14 features, out of which 11 are selected as relevant for the purpose of prediction.
The majority of these features involve measurements related to environmental conditions,
such as $X_{\mathtt{month}}$, $X_{\mathtt{hour}}$, $X_{\mathtt{temperature}}$, $X_{\mathtt{humidity}}$ and $X_{\mathtt{windspeed}}$.
Additionally, certain features provide information about the type of day, for example, whether it is a working day ($X_{\mathtt{workingday}}$) or not.
The target value \( Y_{\mathtt{count}}\) is the bike rentals per hour, which has mean value
\(\mu_{\mathtt{count}} = 189\) and standard deviation \(\sigma_{\mathtt{count}} = 181\).

% Training - Evaluation - Region Extraction
As a black-box model, we train for \(60\) epochs a fully-connected Neural Network with 6 hidden layers, using the Adam optimizer with a learning rate of $0.001$.
The model attains a root mean squared error of \( 0.39 \cdot 181 \approx 70\) counts on the test set.
Subsequently, we extract the subregions, searching for splits up to a maximum spliting depth of \(T=3\).
Following the postprocessing step, we find that the only split that substantially reduces the level of interactions within the subregions is based on the feature
$X_{\mathtt{hour}}$. This feature is divided into two subgroups: $X_{\mathtt{hour}} | \when{X_{\mathtt{workingday}} \neq 1}$ and $X_{\mathtt{hour}} \when{X_{\mathtt{workingday} = 1}}$.

\begin{figure}[htbp]
    \centering
    \begin{subfigure}{0.32\textwidth}
        \centering
        \includegraphics[width=\textwidth]{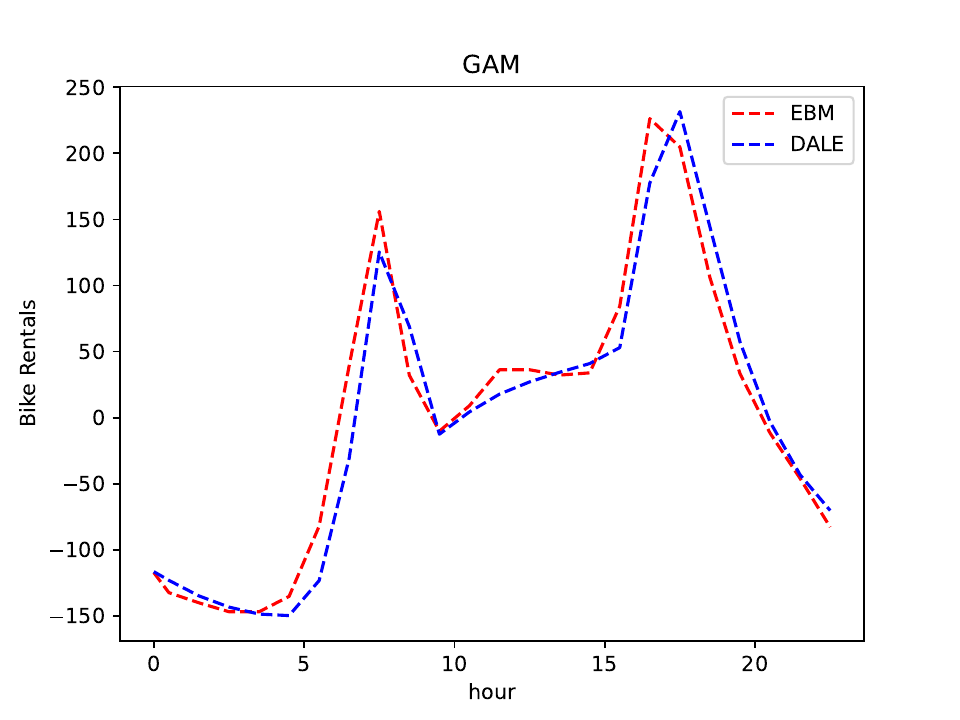}
        \caption{\(f(X_{\mathtt{hour}})\)}
        \label{subfig:bike_rentals_gam}
    \end{subfigure}
    \begin{subfigure}{0.32\textwidth}
        \centering
        \includegraphics[width=\textwidth]{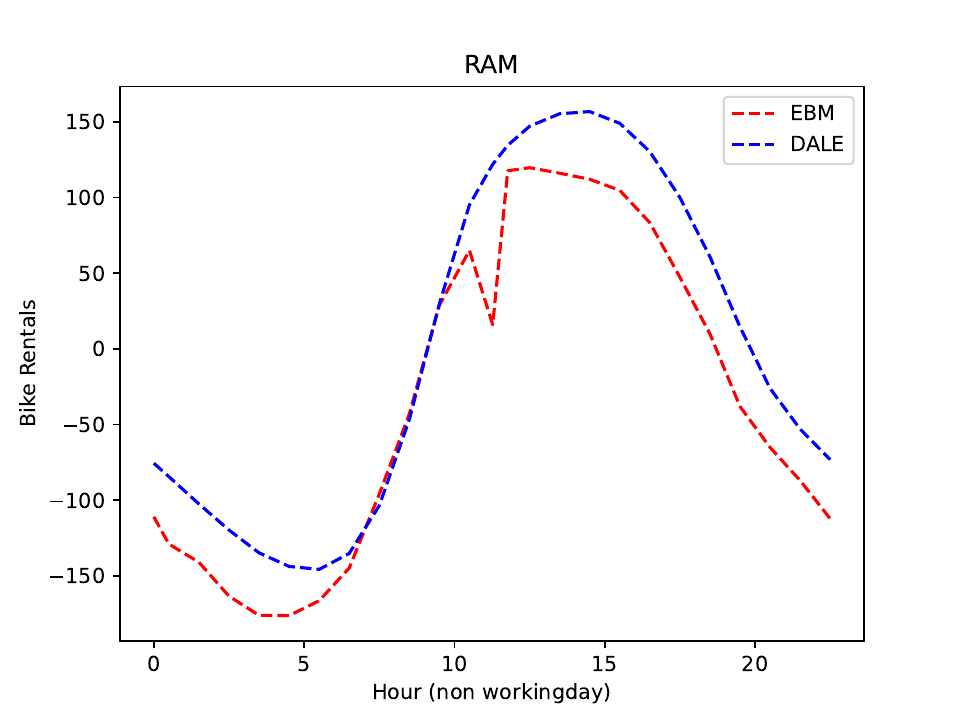}
        \caption{\(f(X_{\mathtt{hour}}) \when{X_{\mathtt{workingday}} \neq 1}\)}
        \label{subfig:bike_rentals_regional_gam_1}
    \end{subfigure}
    \begin{subfigure}{0.32\textwidth}
        \centering
        \includegraphics[width=\textwidth]{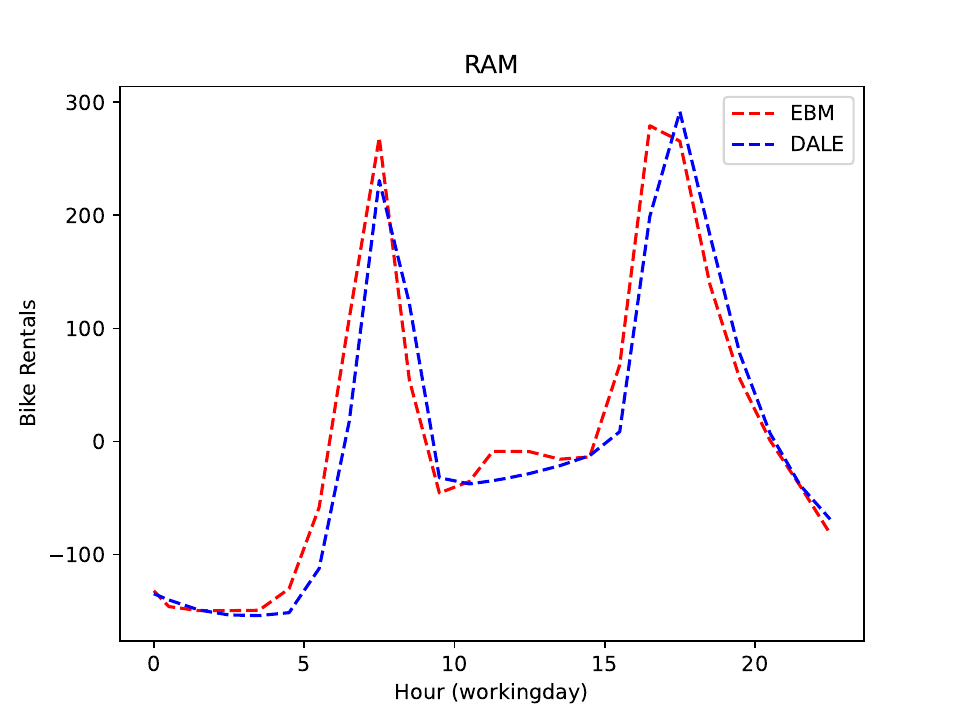}
        \caption{\(f(X_{\mathtt{hour}}) \when{X_{\mathtt{workingday}} = 1}\)}
        \label{subfig:bike_rentals_regional_gam_2}
    \end{subfigure}
    \caption{Comparison of different models' predictions for bike rentals based on the hour of the day.
    Subfigure (a) depicts the generalized additive model (GAM),
        while subfigures (b) and (c) illustrate the RAM model's predictions for different day types: non-working days
        \(f(X_{\mathtt{hour}}) \when{X_{\mathtt{workingday}} \neq 1}\) and
        working days \(f(X_{\mathtt{hour}}) \when{X_{\mathtt{workingday}} = 1}\), respectively.
        The RAM model successfully captures the interaction between the hour of the day and the day type,
        leading to improved predictions and enhanced interpretability.}
    \label{fig:bike_sharing}
\end{figure}

Figure~\ref{fig:bike_sharing} clearly illustrates that the impact of the hour of the day on bike rentals varies
significantly depending on whether it is a working day or a non-working day.
Specifically, during working days, there is higher demand for bike rentals in the morning and afternoon hours,
which aligns with the typical commuting times (Figure~\ref{subfig:bike_rentals_regional_gam_1}).
On the other hand, during non-working days, bike rentals peak in the afternoon as individuals engage in
leisure activities (Figure~\ref{subfig:bike_rentals_regional_gam_2}).
The proposed RAM method effectively captures and detects this interaction by establishing two distinct subregions,
each corresponding to working days and non-working days, respectively.
Subsequently, the EBM that is fitted to each subregion, successfully learns these patterns,
achieving a root mean squared error of approximately \( 0.56 \cdot 181 \approx 101\) counts on the test set.
It is noteworthy that RAM not only preserves the interpretability of the model,
but it also enhances the interpretation of the underlying modeling process.
By identifying and highlighting the interaction between the hour of the day and the day type,
RAM provides valuable insights into the relationship between these variables and their influence on bike rentals.
In contrast, the GAM model~\ref{subfig:bike_rentals_gam} is not able to capture this interaction and
achieves a root mean squared error of \( 0.73 \cdot 181 \approx 132\) counts on the test set.
Finally, in table~\ref{tab:sample}, we also observe that the RA$^2$M, i.e., RAM with second-order interactions,
outperforms the equivalent GA$^2$M model in terms of predictive performance.
Specifically, the RA$^2$M model achieves a root mean squared error of \( 0.41 \cdot 181 \approx 74\) counts,
while the GA$^2$M model of \( 0.44 \cdot 181 \approx 80\) counts on the test set.
It is worth noticing that the RA$^2$M model's accuracy is comparable to the black-box model's accuracy.

\paragraph{California Housing Dataset}

The California Housing dataset consists of approximately $20,000$ of housing blocks situated in California.
Each housing block is described by eight numerical features, namely,
$X_{\mathtt{lat}}$, $X_{\mathtt{long}}$, $X_{\mathtt{median\_age}}$, $X_{\mathtt{total\_rooms}}$, $X_{\mathtt{total\_bedrooms}}$, $X_{\mathtt{population}}$, $X_{\mathtt{households}}$, and $X_{\mathtt{median\_income}}$.
The target variable, $Y_{\mathtt{value}}$, is the median house value in dollars for each block.
The target value ranges in the interval \([15, 500] \cdot 10^3\), with a mean value of
\(\mu_Y \approx 201 \cdot 10^3 \) and a standard deviation of \(\sigma_Y \approx 110 \cdot 10^3\).

\begin{figure}[htbp]
    \centering
    \begin{subfigure}{0.32\textwidth}
        \centering
        \includegraphics[width=\textwidth]{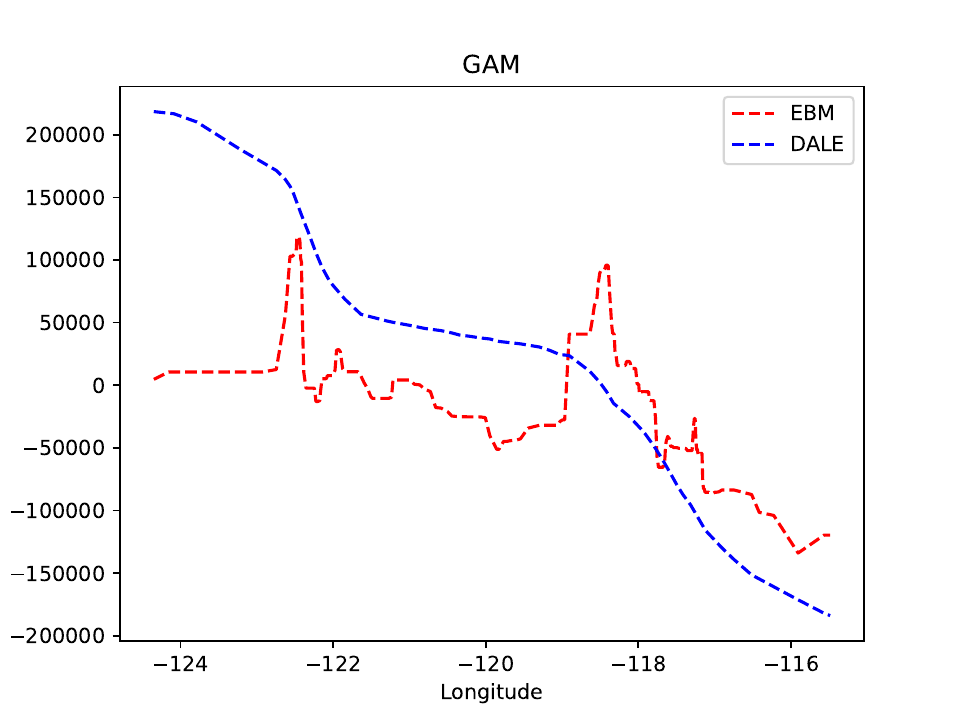}
        \caption{\(f(X_{\mathtt{long}})\)}
        \label{subfig:california_gam}
    \end{subfigure}
    \begin{subfigure}{0.32\textwidth}
        \centering
        \includegraphics[width=\textwidth]{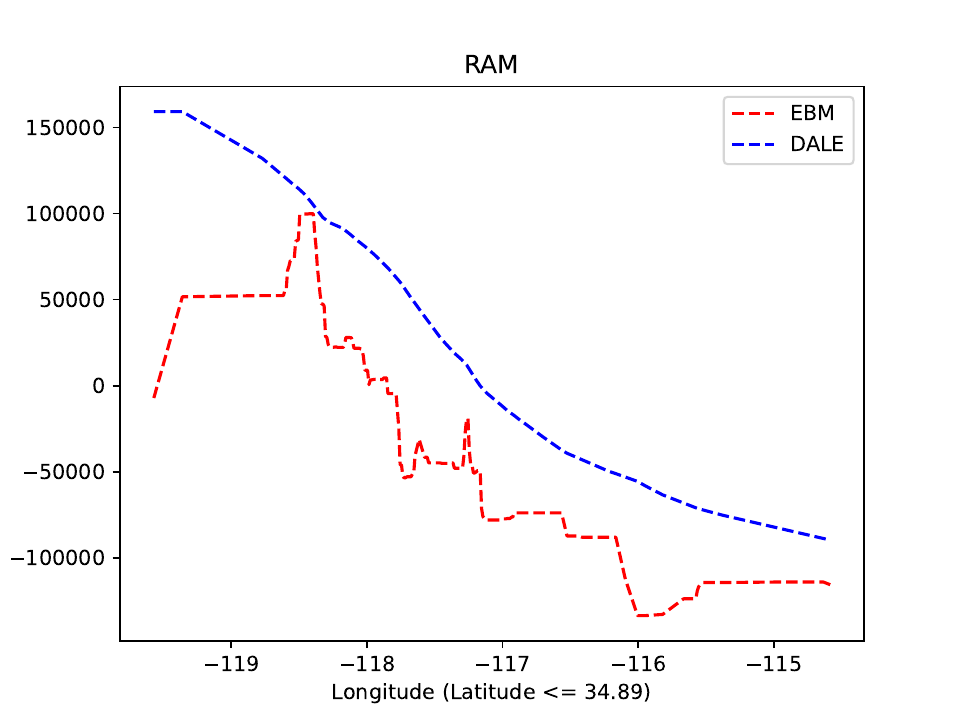}
        \caption{\(f(X_{\mathtt{long}}) \when{X_{\mathtt{lat}} \leq 34.89}\)}
        \label{subfig:california_ram_1}
    \end{subfigure}
    \begin{subfigure}{0.32\textwidth}
        \centering
        \includegraphics[width=\textwidth]{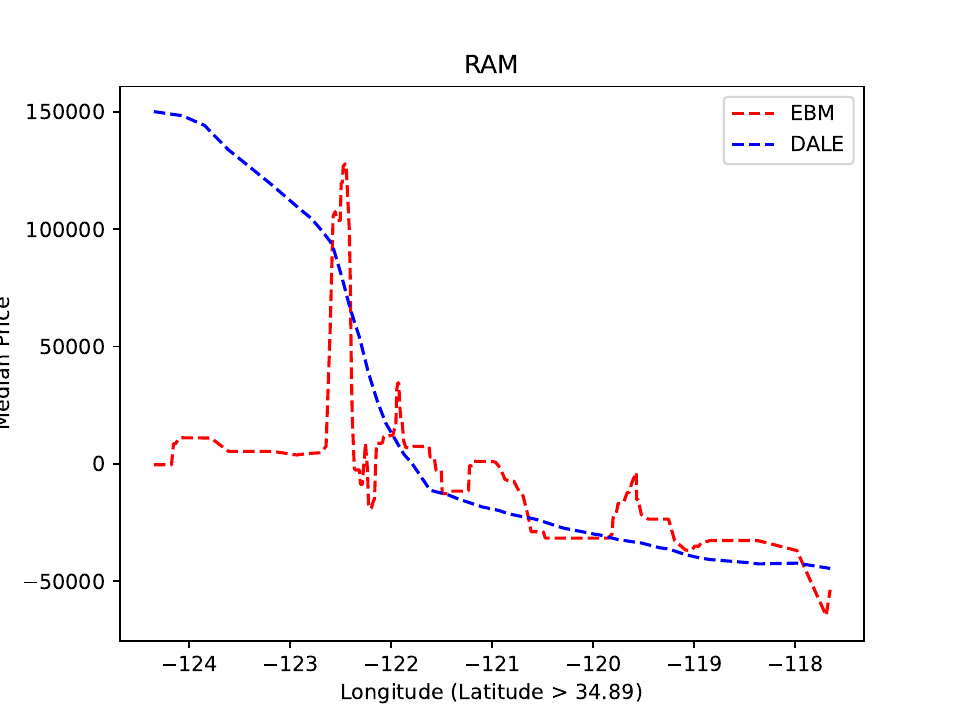}
        \caption{\(f(X_{\mathtt{long}}) \when{X_{\mathtt{lat}} > 34.89}\)}
        \label{subfig:california_ram_2}
    \end{subfigure}
    \caption{Comparison of different predictions for housing prices in California based on the longitude.
    Subfigure (a) showcases the generalized additive model (GAM),
        while subfigures (b) and (c) demonstrate the RAM components for different latitude ranges:
        \(f(X_{\mathtt{long}}) \when{X_{\mathtt{lat}} \leq 34.89}\) and
        \(f(X_{\mathtt{long}}) \when{X_{\mathtt{lat}} > 34.89}\), respectively.
        We observe, that although the EBM model is able to capture the overall trend in the data,
        it also exhibits a large amount of variance.}
    \label{fig:california_housing}
\end{figure}

As a black-box model, we train for $45$ epochs a fully-connected Neural Network with 6 hidden layers,
using the Adam optimizer with a learning rate of $0.001$.
The model achieves a root mean square error (RMSE) of about \(58\)K dollars on the test set.
Subsequently, we perform subregion extraction by searching for splits up to a maximum depth of \(T=3\).
After the postprocessing step, we discover that several splits significantly reduce the level of interactions,
resulting in an expanded input space consisting of \(16\) features, as we show in table~\ref{tab:california_housing_subregions}.
Out of them, we randomly select and illustrate in Figure~\ref{fig:california_housing} the effect of the feature $X_{\mathtt{long}}$.
As we observe, for the house blocks located in the southern part of California ($X_{\mathtt{lat}} \leq 34.9$),
the house value decreases in an almost linear fashion as we move eastward ($X_{\mathtt{long}}$ increases).
In contrast, for the house blocks located in the northern part of California ($X_{\mathtt{lat}} > 34.9$),
the house value performs a rapid (non-linear) decrease as we move eastward ($X_{\mathtt{long}}$ increases).
We also observe that although the EBM fitted to each subregion captures the general trend,
it does not align perfectly with the regional effect.
As in the Bike-Sharing Example, the RMSE of the RAM model, i.e. \( 0.75 \cdot 110 \approx 82.5\)K dollars on the test set,
is lower than the one of the GAM model, i.e.\( 0.82 \cdot 110 \approx 90\)K dollars.
These results indicate that the RAM model provides superior predictions compared to the GAM model.
The same conclusion holds is when comparing the RA$^2$M and the GA$^2$M models,
where the former achieves a RMSE of \( 0.74 \cdot 110 \approx 81\)K dollars,
while the latter of \( 0.77 \cdot 110 \approx 85\)K dollars.

\begin{table}[htbp]
  \centering
  \caption{California Housing: Subregions Detected by RAM}
  \label{tab:california_housing_subregions}
  \begin{tabular}{c|c}
      Feature & Subregions \\
      \hline
      \multirow{2}{*}{$X_{\mathtt{long}}$} & $X_{\mathtt{long}} \when{X_{\mathtt{lat}} \leq 34.9}$ \\
      & $X_{\mathtt{long}} \when{X_{\mathtt{lat}} > 34.9}$ \\
      \hline
      \multirow{2}{*}{$X_{\mathtt{lat}}$} & $X_{\mathtt{lat}} \when{X_{\mathtt{long}} \leq -120.31}$ \\
      & $X_{\mathtt{lat}} \when{X_{\mathtt{long}} > -120.31}$ \\
      \hline
      \multirow{2}{*}{$X_{\mathtt{total\_rooms}}$} & $X_{\mathtt{total\_rooms}} \when{X_{\mathtt{total\_bedrooms}} \leq 449.37}$ \\
        & $X_{\mathtt{total\_rooms}} \when{X_{\mathtt{total\_bedrooms}} > 449.37}$ \\
      \hline
      \multirow{4}{*}{$X_{\mathtt{total\_bedrooms}}$} & $X_{\mathtt{total\_bedrooms}} \when{X_{\mathtt{households}} \leq 411} \when{X_{\mathtt{total\_bedrooms}} \leq 647}$ \\
        & $X_{\mathtt{total\_bedrooms}} \when{X_{\mathtt{households}} \leq 411} \when{X_{\mathtt{total\_bedrooms}} > 647}$ \\
        & $X_{\mathtt{total\_bedrooms}} \when{X_{\mathtt{households}} > 411} \when{X_{\mathtt{total\_bedrooms}} \leq 647}$ \\
        & $X_{\mathtt{total\_bedrooms}} \when{X_{\mathtt{households}} > 411} \when{X_{\mathtt{total\_bedrooms}} > 647}$ \\
      \hline
      \multirow{2}{*}{$X_{\mathtt{population}}$} & $X_{\mathtt{population}} \when{X_{\mathtt{households}} \leq 411.5}$ \\
      & $X_{\mathtt{population}} \when{X_{\mathtt{households}} > 411.5}$ \\
      \hline
      \multirow{2}{*}{$X_{\mathtt{households}}$} & $X_{\mathtt{households}} \when{X_{\mathtt{total\_bedrooms}} \leq 630.57}$ \\
        & $X_{\mathtt{households}} \when{X_{\mathtt{total\_bedrooms}} > 630.57}$ \\
  \end{tabular}
\end{table}

\section{Conclusion and Future Work}

In this paper we have introduced the Regional Additive Models (RAM) framework, a novel approach for learning accurate
x-by-design models from data.
RAMs operate by decomposing the data into subregions, where the relationship between the target variable and the
features exhibits an approximately additive nature.
Subsequently, Generalized Additive Models (GAMs) are fitted to each subregion and combined to create the final model.
Our experiments on two standard regression datasets have shown promising results, indicating that RAMs can provide more accurate predictions compared to GAMs while maintaining the same level of interpretability.

Nevertheless, there are still several unresolved questions that require attention and further experimentation.
Firstly, it is essential to systematically evaluate the performance of RAMs on a larger set of datasets to ensure that the observed improvements are not specific to particular datasets.
Secondly, we need to explore different approaches for each step of the RAM framework.
For the initial step, we should experiment with various black-box models.
Regarding the subregion detection step, we can explore alternative clustering algorithms.
Finally, in the last step, we should investigate different types of GAM models to fit within each subregion.

Another important area of investigation involves exploring the impact of second-order effects within the RAM framework.
While our experimenation demonstrated that even with the current subregion detection, RA$^2$Ms outperform GA$^2$Ms,
it may be the case, that for second-order models the optimal subregions are not necessarily those that maximize the additive effect of individual features,
but rather those that maximize the additive effect of feature pairs.

% \subsubsection{Acknowledgements} Please place your acknowledgments at
% the end of the paper, preceded by an unnumbered run-in heading (i.e.
% 3rd-level heading).

\bibliography{report}

\end{document}